\documentclass[11pt]{article}

\usepackage{amsmath, amssymb}
\usepackage{graphicx}
\usepackage{natbib}
\usepackage{geometry}
\usepackage{hyperref}
\usepackage{booktabs}
\usepackage{authblk}
\usepackage{caption}
\usepackage{subcaption}  % For subfigures

\title{Interpreting Neural Networks through Mahalanobis Distance}
\author{Alan Oursland}
\affil{\textit{alan.oursland@gmail.com}}
\date{October 2024}

\begin{document}

\maketitle

% Abstract
\begin{abstract}
    This paper introduces a theoretical framework that connects neural network linear layers with the Mahalanobis distance, offering a new perspective on neural network interpretability. While previous studies have explored activation functions primarily for performance optimization, our work interprets these functions through statistical distance measures, a less explored area in neural network research. By establishing this connection, we provide a foundation for developing more interpretable neural network models, which is crucial for applications requiring transparency. Although this work is theoretical and does not include empirical data, the proposed distance-based interpretation has the potential to enhance model robustness, improve generalization, and provide more intuitive explanations of neural network decisions.
\end{abstract}

% Sections
\section{Introduction}

Neural networks have revolutionized machine learning, achieving remarkable success across diverse applications. Central to their efficacy is the use of activation functions, which introduce non-linearity and enable the modeling of complex relationships within data. While Rectified Linear Units (ReLU) have gained prominence due to their simplicity and effectiveness \citep{nair2010rectified}, the exploration of alternative activation functions remains an open and valuable area of research \citep{ramachandran2017searching}.

Neural network units are often viewed as linear separators that define decision boundaries between classes \citep{minsky1969perceptrons} with larger activation values suggesting stronger contributions of features to those decisions. Our work challenges this perspective, exploring how individual neurons can be understood through the lens of statistical distance measures. Clustering techniques use distance measures. They aim to minimize the distance between data points and feature prototypes, with smaller values indicating stronger membership to the feature or cluster \citep{macqueen1967methods}. We explore the intersection between these perspectives on activation interpretations, leveraging the distance-minimization approach of clustering techniques to lay the groundwork for novel neural network designs based on statistical distance measures.

This paper establishes a novel connection between neural network architectures and the Mahalanobis distance, a statistical measure that accounts for the covariance structure of data \citep{mahalanobis1936generalized}. We present a robust mathematical framework that bridges neural networks with this statistical distance measure and lays the groundwork for future research into neural network interpretability and design. Our key contributions are:

\begin{enumerate}
    \item We establish a mathematical connection between neural network linear layers and the Mahalanobis distance, demonstrating how Absolute Value (Abs) activations facilitate distance-based interpretations.
    \item We analyze the solution space that neural networks are likely to learn when approximating Mahalanobis distance, exploring the effects of non-uniqueness in whitening transformations and the role of Abs-activated linear nodes.
    \item We discuss the broader implications of this framework for neural network design and interpretability, laying the groundwork for more interpretable models.
\end{enumerate}

% background.tex
\section{Background and Related Work}

\subsection{Activation Functions}

Activation functions introduce non-linearity in neural networks, enabling them to model complex data relationships. The field has evolved from early sigmoid and hyperbolic tangent functions \citep{rosenblatt1958perceptron} to the widely adopted Rectified Linear Unit (ReLU) \citep{nair2010rectified}, which mitigates the vanishing gradient problem in deep networks \citep{glorot2010understanding, krizhevsky2012imagenet}.

ReLU variants that address its shortcomings include Leaky ReLU \citep{maas2013rectifier}, Parametric ReLU (PReLU) \citep{he2015delving}, and Exponential Linear Unit (ELU) \citep{clevert2015fast}. Additionally, newer activation functions like Swish \citep{ramachandran2017searching} and GELU \citep{hendrycks2016gaussian} have been proposed to further enhance network performance and training dynamics.

Tanh and Sigmoid activations are still useful in many architectures such as recurrent neural networks (RNNs) and Long Short-Term Memory (LSTM) networks \citep{hochreiter1997long}.

The variety of activation functions used in modern networks reflects the diverse needs of different architectures. The exploration of activation functions remains an active area of research, with ongoing investigations into their impact on neural network performance, generalization, and interpretability \citep{ramachandran2017searching}. Despite extensive research, the interpretation of activation functions in terms of statistical measures remains an open area of investigation.

\subsection{Overview of Distance Metrics in Clustering}

Distance metrics are fundamental in clustering algorithms, determining how similarity between data points is measured. Various clustering methods employ different distance measures:

\begin{itemize}
    \item K-Means typically uses Euclidean distance ($\ell_2$ norm), assuming spherical clusters and equal feature importance \citep{macqueen1967some}.
    
    \item Gaussian Mixture Models (GMMs) employ Mahalanobis distance, accounting for data covariance and modeling elliptical clusters \citep{reynolds2009gaussian}.

    \item Radial Basis Function (RBF) networks use Gaussian-like activations based on Euclidean distance, creating spherical clusters around learned centers \citep{broomhead1988radial}.

    \item Hierarchical and Agglomerative Clustering can use various metrics (Euclidean, Manhattan, correlation-based), affecting dendrogram shape \citep{murtagh1983survey}.
    
    \item DBSCAN, while often using Euclidean distance, can employ any metric for density-based clustering \citep{ester1996density}.
    
    \item Spectral Clustering incorporates similarity measures like Gaussian kernel functions \citep{von2007tutorial}.
\end{itemize}

The Mahalanobis distance stands out for its ability to account for feature correlations and scale differences, making it particularly useful in multivariate analysis \citep{mahalanobis1936generalized, demaesschalck2000mahalanobis}. It provides a scale-invariant measure that adjusts for the covariance structure of the data, offering advantages in high-dimensional spaces.

Understanding these distance metrics and their properties is crucial for selecting appropriate clustering algorithms and interpreting their results. As we explore the connection between neural networks and distance-based interpretations, these insights from clustering algorithms provide valuable context and inspiration.

\subsection{Neural Network Interpretability and Statistical Models}

The interpretability of neural networks remains a critical challenge, often referred to as the "black-box" nature of these models \citep{lipton2016mythos}. In applications requiring transparency, such as healthcare and finance, understanding the decision-making processes of neural networks is paramount \citep{rudin2019stop}. Various approaches have been developed to enhance interpretability, including feature visualization, saliency maps, and prototype-based methods \citep{erhan2009visualizing, simonyan2013deep, kim2016interpreting}.

Recent advancements in explainable AI (XAI) have introduced tools like SHAP (SHapley Additive exPlanations) \citep{lundberg2017unified} and LIME (Local Interpretable Model-Agnostic Explanations) \citep{ribeiro2016should}, which provide both local and global insights into model predictions. SHAP offers a consistent approach to feature attribution grounded in cooperative game theory, providing robust explanations even for complex models like deep neural networks \citep{lundberg2017unified,markov2020shap}. LIME, by contrast, offers localized interpretability by approximating the model’s behavior with simpler, interpretable models, making it particularly useful for individual predictions \citep{ribeiro2016should,ali2024explainable}.

Connections between neural networks and statistical models, such as Bayesian neural networks \citep{neal1996bayesian,blundell2015weight}, continue to provide a probabilistic framework for understanding model uncertainty. Moreover, concept-based interpretability methods like TCAV (Testing with Concept Activation Vectors) \citep{kim2018interpretability} have emerged, allowing for a more granular analysis of what neural networks learn, which can be crucial in high-stakes domains like healthcare \citep{hanif2024systematic}.

While significant strides have been made in explaining model behavior, there remains a gap in establishing direct mathematical connections between neural network components, specifically activation functions, and statistical distance measures like the Mahalanobis distance. Addressing this gap can provide deeper insights into feature learning and decision-making processes, enhancing both interpretability and robustness of neural network models.

% math_framework.tex

\section{Mathematical Framework}
\label{sec:math_framework}
Gaussians fall out of second-order Taylor series approximations \citep[Section 4.4]{bishop2006pattern}, making them effective for modeling data, even when the data is not explicitly Gaussian. Gaussian mixtures can serve as piecewise linear approximations of complex distributions and surfaces. They are well-suited for modeling point clouds such as the ones neural networks are trained on.

In this section, we develop the mathematical foundation that connects neural networks to the Mahalanobis distance, thereby providing a framework for interpreting neural network operations through the lens of statistical distance metrics. We begin by revisiting key concepts related to Gaussian distributions and the Mahalanobis distance, followed by a detailed exploration of how neural network components, particularly linear layers and activation functions, can approximate these distance metrics. This framework not only enhances our understanding of neural network behavior but also lays the groundwork for leveraging statistical principles to improve network interpretability and training dynamics.

\subsection{Mahalanobis Distance for a Multivariate Gaussian Distribution}

A multivariate Gaussian (Normal) distribution is a fundamental concept in statistics, describing a \(d\)-dimensional random vector \(\mathbf{x} \in \mathbb{R}^d\) with a mean vector \(\boldsymbol{\mu} \in \mathbb{R}^d\) and a covariance matrix \(\boldsymbol{\Sigma} \in \mathbb{R}^{d \times d}\) \citep{bishop2006pattern}. We denote this distribution as \(\mathbf{x} \sim N(\boldsymbol{\mu}, \boldsymbol{\Sigma})\).

The Mahalanobis distance quantifies the distance between a point \(\mathbf{x}\) and the mean \(\boldsymbol{\mu}\) of a distribution, while considering the covariance structure of the data \citep{mahalanobis1936generalized, demaesschalck2000mahalanobis}. It is defined as:

\begin{equation}
\label{eq:mahalanobis_distance}
D_M(\mathbf{x}) = \sqrt{ (\mathbf{x} - \boldsymbol{\mu})^\top \boldsymbol{\Sigma}^{-1} (\mathbf{x} - \boldsymbol{\mu}) }.
\end{equation}

This metric adjusts for variance across dimensions by effectively whitening the data, resulting in a spherical distance measure.

\subsection{Principal Component Analysis (PCA)}

Principal Component Analysis (PCA) is a dimensionality reduction technique that transforms data into a new coordinate system, emphasizing directions (principal components) that capture the most variance \citep{jolliffe2002principal}. When performing PCA on the covariance matrix $\boldsymbol{\Sigma}$, it is decomposed using eigenvalue decomposition:

\begin{equation}
    \label{eq:pca_decomposition}
    \boldsymbol{\Sigma} = \mathbf{V} \boldsymbol{\Lambda} \mathbf{V}^\top,
\end{equation}

where:
\begin{itemize}
    \item $\mathbf{V} = [\mathbf{v}_1, \mathbf{v}_2, \dots, \mathbf{v}_d]$ is a matrix whose columns are the orthogonal unit eigenvectors of $\boldsymbol{\Sigma}$.
    \item $\boldsymbol{\Lambda} = \text{diag}(\lambda_1, \lambda_2, \dots, \lambda_d)$ is a diagonal matrix of the corresponding eigenvalues $\lambda_i$, representing the variance along each principal component.
\end{itemize}

Substituting $\mathbf{V} \boldsymbol{\Lambda} \mathbf{V}^\top$ for $\boldsymbol{\Sigma}$ in the Mahalanobis distance equation \eqref{eq:mahalanobis_distance}, we obtain:

\begin{equation}
    \label{eq:mahalanobis_pca}
    D_M(\mathbf{x}) = \sqrt{ (\mathbf{x} - \boldsymbol{\mu})^\top \mathbf{V} \boldsymbol{\Lambda}^{-1} \mathbf{V}^\top (\mathbf{x} - \boldsymbol{\mu}) }.
\end{equation}

Simplify to express the Mahalanobis distance in terms of individual component contributions:

\begin{align}
D_M(\mathbf{x}) &= \sqrt{ (\mathbf{x} - \boldsymbol{\mu})^\top \mathbf{V} \boldsymbol{\Lambda}^{-1} \mathbf{V}^\top (\mathbf{x} - \boldsymbol{\mu}) } \nonumber \\
&= \sqrt{ (\mathbf{V}^\top (\mathbf{x} - \boldsymbol{\mu}))^\top \boldsymbol{\Lambda}^{-1} (\mathbf{V}^\top (\mathbf{x} - \boldsymbol{\mu})) } \nonumber \\
&= \sqrt{ \sum_{i=1}^{d} \lambda_i^{-1} \left( \mathbf{v}_i^\top (\mathbf{x} - \boldsymbol{\mu}) \right)^2 } \nonumber \\
&= \left\| \lambda_i^{-1/2} \mathbf{v}_i^\top (\mathbf{x} - \boldsymbol{\mu}) \right\|_2.
\label{eq:mahalanobis_pca_l2}
\end{align}

where $\| \cdot \|_2$ denotes the Euclidean ($\ell_2$) norm. This shows that the Mahalanobis distance can also be expressed as the $\ell_2$ norm of the number of standard deviations of \(\mathbf{x}\) along each principal component.

\subsection{Connecting Neural Networks to Mahalanobis Distance}

We consider the Mahalanobis distance along a single principal component.

\begin{equation}
    \label{eq:mahalanobis_single_component}
    D_{M,i}(\mathbf{x}) = \left| \lambda_i^{-1/2} \mathbf{v}_i^\top (\mathbf{x} - \boldsymbol{\mu}) \right|,
\end{equation}

This equation projects the centered data $(\mathbf{x} - \boldsymbol{\mu})$ onto the direction of variance defined by the principal component eigenvector and scales it by the inverse square root of the eigenvalue.

Let 
\begin{align}
    \mathbf{W} &= \lambda_i^{-1/2} \mathbf{v}_i^\top, \\
    \mathbf{b} &= - \lambda_i^{-1/2} \mathbf{v}_i^\top \boldsymbol{\mu}.
\end{align}

We can simplify Equation \eqref{eq:mahalanobis_single_component} to

\begin{equation}
    \label{eq:mahalanobis_linear}
    D_{M,i}(\mathbf{x}) = \left| \mathbf{W} \mathbf{x} - \mathbf{b} \right|,
\end{equation}

This is identical to the equation for a linear layer where $\boldsymbol{W}$ represents the weight matrix, $\boldsymbol{b}$ the bias vector, and the Abs function serves as the activation function. Each linear node with an Abs activation can be interpreted as modeling a one-dimensional Gaussian along a principal component direction, with the decision boundary passing through the mean of the modeled cluster. The layer as a whole represents a subset of principal components from a Gaussian Mixture Model (GMM) that approximates the input distribution. Since each component captures significant features individually, we do not need to aggregate them via an $\ell_2$ norm (full Mahalanobis distance computation). Instead, the subsequent layer clusters these principal component features, effectively forming a new GMM that models the outputs of the first layer.

\subsection{Non-Uniqueness of Whitening}

The principal components of a Gaussian distribution, as used in the Mahalanobis distance, form an orthonormal set of axes. Projecting Gaussian data onto these axes transforms the distribution from an oriented ellipsoid into a spherical Gaussian, effectively converting the data from \(\mathcal{N}(\boldsymbol{\mu}, \boldsymbol{\Sigma})\) to \(\mathcal{N}(\mathbf{0}, \mathbf{I})\). This process is known as \emph{whitening} \citep[Section 12.1.3]{bishop2006pattern}.

However, the transformation to whitened data is not unique. Specifically, any rotation applied in the whitened space results in another valid whitening transformation. Mathematically, if \(\mathbf{x}_w\) is the whitened data, then for any orthogonal rotation matrix \(\mathbf{R} \in \text{SO}(d)\), the rotated data \(\mathbf{x}_w' = \mathbf{R} \mathbf{x}_w\) is also whitened.

This non-uniqueness implies that multiple sets of axes, possibly non-orthogonal in the original space, can serve as a whitening basis. When we transform the rotated basis back to the original space, we obtain a new set of basis vectors \(\mathbf{W}\) that still whiten the data but may not correspond to the original principal components and may not even be orthogonal.

In the context of neural networks, this means that although linear nodes can represent directions that effectively whiten the data, they are unlikely to precisely learn the actual principal components when estimating Mahalanobis distances. Instead, they may learn any basis that achieves whitening. Nevertheless, the learned hyperplanes (decision boundaries) should still pass through the data mean \(\boldsymbol{\mu}\), allowing for prototype interpretation.

To encourage the network to learn the actual principal components, one could apply an orthogonality constraint or regularization on the weight matrices. This regularization promotes learning orthogonal directions, aligning the learned basis with the true principal components of the data clusters and providing statistically independent features.

% discussion.tex

\section{Implications and Discussion}
\label{sec:discussion}

We discuss implications, potential impact and future work of this reframing of linear layers in neural networks. While this paper provides a robust theoretical foundation for interpreting neural networks through Mahalanobis distance and Abs activation functions, it does not include empirical results. Future work will involve validating these theoretical insights with empirical data to further assess their applicability and performance in real-world scenarios.

\subsection{Expected Value Interpretation}

The expected value, or mean, is a central concept in statistics, representing the average tendency of a distribution. In neural networks, finding the expected value for each neuron would reveal the features it recognizes. Interpreting linear nodes as approximations of Gaussian principal components  provides a path towards recovering the neuron mean value. The estimated mean serves as a prototype for the feature that the neuron has learned to recognize \citep{li2018deep}, representing the 'ideal' input for that neuron. This interpretation enhances the transparency of the feature extraction process, potentially leading to more interpretable models and improved architectures.

\subsection{Equivalence between Abs and ReLU Activations}

While our analysis utilizes linear layers with Abs activation functions to model deviations along principal component directions, ReLU activations can provide comparable information within the same framework.

For the \emph{Abs activation}, each linear node computes:
\begin{equation}
\label{eq:abs_activation_compact}
y_{\text{Abs}} = \left| \mathbf{w}^\top \mathbf{x} + b \right|,
\end{equation}
where the weights \(\mathbf{w}\) and bias \(b\) are set such that \(\mathbf{w}^\top \boldsymbol{\mu} + b = 0\). This centers the decision boundary at the cluster mean \(\boldsymbol{\mu}\), and within a confidence interval \(\delta\), the pre-activation output ranges from \(-\delta\) to \(+\delta\).

For the \emph{ReLU activation}, we adjust the bias to shift the decision boundary just outside the cluster:
\begin{equation}
\label{eq:relu_activation_compact}
y_{\text{ReLU}} = \max\left( 0, -\mathbf{w}^\top \mathbf{x} - b + \delta \right).
\end{equation}
Here, the pre-activation output ranges from \(0\) to \(2\delta\) within the cluster. Although ReLU zeros out negative inputs, by negating the pre-activation and adjusting the bias, it effectively captures the magnitude of deviations similar to the Abs activation.

The hyperplanes defined by \(\mathbf{w}\) maintain the same orientation in both cases, providing equivalent views of the cluster. Subsequent layers can adapt to either activation's output range, making Abs and ReLU functionally comparable in capturing essential features.

This suggests that techniques developed for networks with Abs activations may be adaptable to ReLU activations, bridging theoretical insights with practical neural architectures commonly utilizing ReLU.

\subsection{Activations as Distance Metrics}

Traditional neural networks typically employ an ``intensity metric model,'' where larger activation values indicate stronger feature presence. In contrast, a ``distance metric model'' interprets smaller activation values as indicating closer proximity to a learned feature or prototype. The following observations suggest directions for future work:

\begin{itemize}
    \item Most error functions (e.g., Cross Entropy Loss, Hinge Loss) are designed for intensity metrics. Output layers using Abs activation may require modification of their output values.
    \item While some architectures, like Radial Basis Function networks \citep{broomhead1988radial}, utilize distance metrics, they are not widely adopted in modern deep learning.
    \item Distance metrics conflict with the goal of sparse output layers. In a distance metric model, zero is the strongest signal, making it illogical for most outputs to have the strongest signal.
    \item The Gaussian connection suggests transforming distance metrics through exponential ($y=e^{-x^2}$) or Laplace ($y=e^{-|x|}$) functions to convert them into intensity metrics. However, these may suffer from vanishing gradients. A approximation of these functions could combine Abs and ReLU: $y=\text{ReLU}(-\text{Abs}(x) + \text{confidence\_bound})$.
    \item Distance and intensity metrics can be interconverted through negation. Subsequent layer weights can apply their own negation, obscuring the metric type learned by internal nodes.
    \item There may exist regularization techniques that encourage distance metric learning \citep{weinberger2009distance}.
\end{itemize}

\subsection{Model Initialization and Pretraining}

Interpreting neurons as learning distances from cluster means suggests novel approaches to model initialization and pretraining. This perspective offers an alternative to standard random initialization techniques \citep{kamilov2017survey} by incorporating data-driven insights into the model's starting configuration.

Rather than initializing with random weights, an approach could involve clustering the input data (e.g., using k-means) and calculating the covariance of each cluster. Applying Principal Component Analysis (PCA) to these covariance matrices can provide a basis for directly initializing network parameters. This strategy leverages the structure of the data to guide the network's early learning stages. This approach and its approximations may offer several advantages:

\begin{itemize} \item Faster convergence by starting with parameters informed by the data distribution \item Enhanced interpretability, as network weights are aligned with meaningful features from the outset \item Improved generalization by incorporating information about cluster structures \end{itemize}

\subsection{Model Translation and Componentization}

The interpretation of neurons as principal components of Gaussians suggests a potential mapping between neural networks and hierarchical Gaussian Mixture Models (GMMs) \citep{jacobs1991adaptive}. By performing PCA on the clusters in a GMM, we can extract principal components, converting them directly into neurons. Conversion from neurons to Gaussian representations may also be possible. The process of directly translating between neural networks and GMMs offers several potential advantages:

\begin{itemize}
    \item \textbf{Enhanced Interpretability}: Neural networks can be better understood through their GMM equivalents, providing insights into the data distribution and feature representations.
    \item \textbf{Application of Statistical Techniques}: Established statistical methods used in GMM analysis can be applied to neural networks, potentially improving training and evaluation.
    \item \textbf{Hybrid Models}: Combining neural networks and GMMs can leverage the strengths of both, enhancing performance in tasks like clustering and classification.
    \item \textbf{Model Decomposition}: Large networks might be decomposable into smaller, context-specific subnetworks, facilitating easier analysis and maintenance.
    \item \textbf{Efficient Storage and Computation}: Subnetworks can be stored offline and dynamically loaded based on data context, improving memory efficiency and reducing computational overhead.
    \item \textbf{Scalability in Large-Scale Applications}: This approach can lead to faster inference and more efficient resource utilization in applications dealing with massive datasets.
\end{itemize}

\subsection{Direct use of Mahalanobis equation}

Equation \ref{eq:mahalanobis_single_component} explicitly incorporates the variance eigenvalue $\lambda$, the unit eigenvector $\mathbf{v}$, and the mean $\boldsymbol{\mu}$. Batch Normalization already makes use of $\lambda$ and $\boldsymbol{\mu}$ \citep{ioffe2015batch}, while the nGPT model employs unit weight vectors, which are analogous to $\mathbf{v}$ \citep{loshchilov2024ngptnormalizedtransformerrepresentation}. The success of these techniques suggests there might be further opportunities to decompose the standard linear layer equation $y = Wx + b$ towards the Mahalanobis equation in a way that leads to improvements in training speed and representation quality.

% conclusions.tex

\section{Conclusion}
\label{sec:conclusion}

This paper establishes a novel connection between neural network architectures and the Mahalanobis distance, providing a fresh perspective on neural network interpretability. By demonstrating how linear layers with Abs activations can approximate Mahalanobis distances, we bridge the gap between statistical distance measures and neural network operations. This framework offers several key insights:

\begin{itemize}
    \item It provides a probabilistic interpretation of neural network nodes as learning principal components of Gaussian distributions.
    \item It suggests new approaches for model initialization, pretraining, and componentization.
    \item It establishes a potential homomorphism between neural networks and hierarchical Gaussian Mixture Models.
\end{itemize}

These findings lay the groundwork for future research into more interpretable and robust neural network architectures. By leveraging statistical principles in neural network design, we open new avenues for enhancing model transparency, improving generalization, and developing more efficient training techniques. As the field of AI continues to evolve, such interpretable frameworks will be crucial in building trustworthy and explainable AI systems.

\bibliographystyle{plainnat}
\bibliography{references}

\end{document}